\documentclass{svproc}

\usepackage{float}
\usepackage{amssymb}
\usepackage{amsmath}
\usepackage{amsfonts}
\usepackage{amstext}
\usepackage{tabularx}
\usepackage{algorithm}
\usepackage{algpseudocode}

\usepackage{cite}
\usepackage{latexsym}
\usepackage{booktabs}
\usepackage{enumitem}
\usepackage{color}

\usepackage{textcomp}
\usepackage{multirow}
\usepackage{multicol}
\usepackage{siunitx}
\usepackage[table]{xcolor}
\usepackage{graphicx}
\usepackage{hyperref}
\usepackage{subfig}
\usepackage[symbol]{footmisc}

\begin{document}
\mainmatter              
\title{Change Point Detection with Generative Models}
\titlerunning{Change Point Detection with Generative Models}  
%
\author{Tigran Ramazyan\inst{1} \and Denis Derkach\inst{1} \and Mikhail Hushchyn\inst{1}}
\authorrunning{Tigran Ramazyan et al.} 
%
\tocauthor{Tigran Ramazyan, Denis Derkach, Mikhail Hushchyn}
\institute{HSE University, 11 Pokrovsky Boulevard, Moscow, 109028, Russia. \\
e-mail: \email{tramazyan@hse.ru}\\[4pt]
  {\small Received: August 11, 2026; revised: August 20, 2026; accepted: August 31, 2026.}
}

\maketitle              

\begin{abstract}
Change point detection (CPD) methods aim to identify abrupt changes in the properties of time series, which is a key problem in numerous fields. A major class of CPD approaches relies on computing a CPD score, which is either a discrepancy or divergence between two subsequent samples, called windows. A fundamental limitation of window-based CPD is the 'sample size bottleneck.' To ensure temporal precision, windows must be small; however, small samples provide insufficient statistical evidence for the classifiers and deep-learning models currently used to compute CPD scores. This leads to high variance in the scores and frequent mispredictions. Our study tackles this issue by exploring generative models for window augmentation, effectively expanding available data to improve CPD quality. Experiments on synthetic and real data demonstrate that the proposed approach significantly enhances CPD. Specifically, we show that this approach allows simple classifiers to outperform complex deep-learning baselines by stabilizing the CPD score distribution.

Source code available via the \href{https://github.com/ramazyant/GMCPD}{link}.
\end{abstract}
\section{Introduction}\label{sec:intro}
Change point detection (CPD) is a crucial tool in the analysis of time series data, allowing researchers and practitioners to identify abrupt changes in the behaviour of a system. Change points often signal significant critical events, which makes CPD widely applicable in various domains~\cite{tartakovsky01151824}. Most notably on ECG data for healthcare monitoring~\cite{10.5555/151741}, on seismic data to predict natural disasters~\cite{Brodsky_1993}, on network data to prevent cybersecurity breaches~\cite{Csorgo1997LimitTI}, and on stock price data for investing decisions~\cite{9741807}. Each of these applications depends on the accurate CPD for a timely response to a critical event.\newline \indent
The first change point detection algorithms were developed in the 1950s for quality control of manufacturing processes~\cite{10.1214/aoms/1177729489}. Since then, the task of detecting change points has been further investigated, and new statistical and machine-learning methods have been developed. CPD is conducted online or offline, as in the works of~\cite{tartakovsky01151824,10.1145/1553374.1553418} and~\cite{tartakovsky01151824,Moskvina2006ApplicationOT}, respectively. In online settings, a new sample instance is provided in real-time. It is essential predict change points with minimal delay while maintaining a low misprediction rate. The entire sample is provided in offline settings. The goal in this case is to achieve the best prediction quality. In this study, all algorithms were used in offline settings.\newline \indent
The most common approach for CPD is to compare the densities of the observations in two subsequent samples, so-called windows. The cumulative sum (CUSUM) method developed assumes the density in each window and examines each observation by performing sequential hypothesis tests~\cite{10.5555/151741}. Similar to CUSUM, the generalised likelihood ratio (GLR) approach in~\cite{tartakovsky01151824} assumes the density in each window and then estimates the distribution parameters by maximising the likelihood. \newline \indent
Furthermore, over the last decade, several cost function methods have been proposed ~\cite{TRUONG2020107299}. The cost functions encapsulate the nature of the change points of interest, and are treated as discrepancy measures. Cost-function algorithms also differ in their CPD approaches. For example, Window by~\cite{TRUONG2020107299} computes the discrepancy between two windows throughout the timeline. The binary segmentation algorithm (BinSeg) by~\cite{10.1214/14AOS1245} splits the sample into two at points of the highest cost function value and proceeds iteratively on each subsample. While our work focuses on window-based discrepancy, other classic paradigms include search-based methods like PELT \cite{TRUONG2020107299} and Bayesian approaches, see references \cite{Adams2007, tsaknaki2025bayesian}.\newline \indent
As machine learning methods have been developed, starting with the work of~\cite{sugiyama_suzuki_kanamori_2012}, several CPD methods have been proposed that directly estimate the probability density ratio of the two windows. These methods include logistic regression on RBF kernels by~\cite{10.1145/1273496.1273507} and other RBF kernel methods such as KLIEP, uLSIF and RuLSIF ~\cite{NIPS2007_be83ab3e, JMLR:v10:kanamori09a}, and~\cite{10.1162/NECO_a_00442}. In a comparative study, ~\cite{10.1007/s10115-016-0987-z} showed that RuLSIF was the best among these methods.\newline \indent
Artificial neural networks have induced a breakthrough in high-dimensional data analysis \cite{xu2025change}, especially for image, video, sound generation and recognition.~\cite{chang2019kernel} and~\cite{deryck2021change} presented KL-CPD and TIRE respectively. These methods, along with~\cite{HUSHCHYN2021101385} and ~\cite{Ryzhikov_2023}, adopt the deep learning paradigm and leverage it to outperform a wide range of baselines.\newline \indent
With generative modelling being one of the hottest deep-learning topics in the last decade, it has found its place within the CPD problem. For example, \cite{kei2025changepointdetectiondynamic} suggested a simple generative decoder-only architecture to learn priors for graph embedding. The change points were identified by detecting the prior distribution shifts. Similarly, \cite{KENDRICK20181} used generative priors to adapt to evolving graph structures, thereby improving detection in social networks and IoT systems.\newline \indent
The remainder of this paper is organised as follows. In Section~\ref{sec:prelim} we provide the preliminaries for our study. Section~\ref{sec:method} presents the proposed method and provides theoretical guarantees. Section~\ref{sec:exp} details the datasets, quality metrics, and baselines used in the experiments. In Section~\ref{sec:results} we demonstrate that generative data augmentation significantly enhances the accuracy and robustness the of CPD algorithms. The research findings are discussed in Section~\ref{sec:discussion}, and conclusions are presented in Section~\ref{sec:conclusion}.

\section{Preliminaries}\label{sec:prelim}
Consider $N$ variables of a $d$-dimensional time series:
\begin{equation}
    X = [x_1, x_2, \ldots, x_N] \in \mathbb{R}^{d \times N},
\end{equation}
where $x_t$ comes at time $t$. Assuming that there is a change point at an unknown time $\nu \geq 1$, without loss of generality, we can describe the change point as follows. Before the change point, data follow density $p_0: x_1, x_2, \ldots, x_{\nu} \sim p_0(x_i), \forall i \leq \nu$ and another density after the change point: $x_{\nu + 1}, x_{\nu + 2}, \ldots, x_{n} \sim p_1(x_i), \forall i > \nu$.\newline \indent
We begin with a general window-based CPD algorithm. This process is presented in Algorithm~\ref{alg:cpd}. The algorithm proceeds iteratively through observations. At each time $t$, the original data are first divided into pairs of reference and test windows, $X_{ref}$ and $X_{test}$. Both windows are Hankel matrices from $n$ subsequent observations. The CPD score is then estimated, which is either distance or divergence. When neither $X_{ref}$ nor $X_{test}$ contains a change point, $D$ is usually infinitesimal and may contain minor fluctuations. As the windows moved along the data, $X_{test}$ was the first to capture a change point, significantly increasing the CPD score. Because $D$ is either a distance or divergence, it reaches a maximum value when the change point is exactly between $X_{ref}$ and $X_{test}$. Subsequently, the CPD score decreased. Thus, predicting the change points with any window-base algorithm boils down to finding peaks, that is, local maxima in the CPD score.

\begin{algorithm}[ht!] 
\caption{Window-based CPD}
\label{alg:cpd}
\begin{algorithmic}[1]
\Require{CPD score $D$, data $X \in \mathbb{R}^{d \times N}$,  window size $n$, peak threshold $\tau$}

\State Initiate CPD score estimates: $\hat{D} \gets \varnothing$
\For{$t = n \dots N - n$ do}
    \State $X_{ref} \gets X_{t - n + 1 : t}$;
    \State $X_{test} \gets X_{t + 1 : t + n}$;
    \State Compute CPD score at $t$: $d_t \gets D(X_{ref}, X_{test})$;
    \State Update CPD score estimates: $\hat{D} \gets \hat{D} \bigcup \{d_t\}$;
\EndFor
\State $\hat{T} \leftarrow \{t \mid d_t \text{ is a local maximum and } d_t > \tau\}$;
\State \Return $\hat{T}$.
\end{algorithmic}
\end{algorithm}

Algorithms such as Window~\cite{TRUONG2020107299} use metric-based CPD scores, such as, the $L_2$ score. Many other methods rely on divergences; for example, RuLSIF ~\cite{LIU201372} uses symmetric Kullback-Leibler (KL) divergence as the CPD score. Kernel-based methods like Maximum Mean Discrepancy (MMD) offer an alternative to KL-divergence but often struggle with high-dimensional variance in small-sample regimes. The divergences were computed between $P_{ref}, X_{ref} \sim P_{ref}$ and $P_{test}, X_{test} \sim P_{test}$. However, these densities are typically unknown. Modelling densities is a difficult task, and incorporating density estimation into a CPD algorithm may further complicate the task.

However, only the density ratio of $P_{ref}$ and $P_{test}$ is required to compute the CPD score. As argued in~\cite{sugiyama_suzuki_kanamori_2012} and ~\cite{10.1145/1273496.1273507}, directly estimating density ratios from data is suboptimal. They also show the probability density ratio is the output of an ideal probabilistic classifier treated as a discriminator model. Thus, by fitting a probabilistic classifier with $x \in X_{ref}$ labelled as $y=0$ and $x \in X_{test}$ labelled as $y=1$, the density ratio can easily estimated as follows:

\begin{equation}
    \frac{P_{test}}{P_{ref}} = \frac{P(y = 1 | x)}{P(y = 0 | x)} = \frac{P(y = 1 | x)}{1 - P(y = 1 | x)}.
\end{equation}

Take, for example, symmetric KL divergence as the CPD score:

\begin{equation}\label{eq:kl_sym}
    D = KL(P_{ref}||P_{test}) + KL(P_{test}||P_{ref}).
\end{equation}

Given a density ratio estimator, $D$ can be estimated as

\begin{equation}\label{eq:est_kl_sym}
    \begin{split}
        \hat{D} = &\frac{1}{n} \sum\limits_{x \in X_{ref}} \log \frac{P(y=1|x)}{1-P(y=1|x)} +\\ +&\frac{1}{n} \sum\limits_{x \in X_{test}} \log \frac{1-P(y=1|x)}{P(y=1|x)},
    \end{split}
\end{equation}

The resulting score, $\hat{D}$, quantifies the probability that a change point occurs between the reference and test windows, with higher values indicating a higher probability of the presence of a change point.

The authors of~\cite{LIU201372} have shown that using a single forward or reverse divergence may not detect all change points. Logically, the behaviours of forward and reverse divergences as CPD scores are opposite to each other.

A general CPD algorithm with a binary classifier is given in Algorithm~\ref{alg:clf_cpd}.

\begin{algorithm}[ht!] 
\caption{Window-based binary classifier CPD}
\label{alg:clf_cpd}
\begin{algorithmic}[1]
\Require{CPD score $D$, data $X \in \mathbb{R}^{d \times N}$, window size $n$, binary classifier $Clf$, peak threshold $\tau$} 

\State Initiate CPD score estimates: $\hat{D} \gets \varnothing$
\For{$t = n \dots N - n$ do}
    \State $X_{ref} \gets X_{t - n + 1 : t}$;
    \State $X_{test} \gets X_{t + 1 : t + n}$;
    \State $y_{ref} \gets 0, y_{test} \gets 1$;
    \State Fit classifier: $Clf ((X_{ref}, y_{ref}), (X_{test}, y_{test}))$;
    \State Compute CPD score at $t$: $d_t \gets D(Clf, X_{ref}, X_{test})$;
    \State Update CPD score estimates: $\hat{D} \gets \hat{D} \bigcup \{d_t\}$;
\EndFor
\State $\hat{T} \leftarrow \{t \mid d_t \text{ is a local maximum and } d_t > \tau\}$;
\State \Return $\hat{T}$.
\end{algorithmic}
\end{algorithm}

While Algorithm \ref{alg:clf_cpd} provides the standard framework for classifier-based detection, its reliance on the raw window size $n$ leaves it vulnerable to stochastic fluctuations. In the following section, we describe how generative models can be integrated into this loop to artificially expand $n$, thereby stabilizing the score $D$.

\section{Method}\label{sec:method}

In this section, we describe the improvement of CPD algorithms using generative models (GM). By generative modelling, we intend statistical learning models that can  fit over a window of samples, that is, $X_{ref}$ or $X_{test}$, to learn the corresponding underlying distributions $P_{ref}$ and $P_{test}$. Subsequently, by generating synthetic samples for both the reference and test windows, we can increase the statistical evidence available to the CPD algorithm. This process improves the robustness of the CPD algorithm, leading to better detection accuracy, even in challenging scenarios, such as noisy data or subtle changes. From this point on, we exclusively use the symmetric KL, Eq.~\ref{eq:kl_sym}, as the CPD score $D$ and its estimator $\hat{D}$, Eq.~\ref{eq:est_kl_sym}.

\begin{figure*}[ht]
    \centering
    \includegraphics[width=\textwidth]{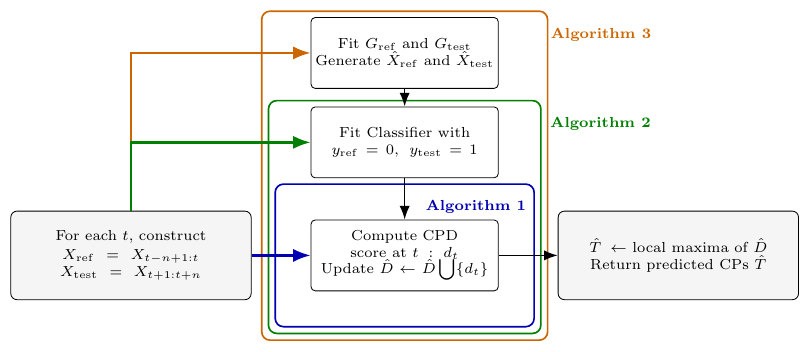}
    \caption{Visual demonstration of Algorithms~\ref{alg:cpd}, ~\ref{alg:clf_cpd}, and ~\ref{alg:gen_cpd}. Algorithm~\ref{alg:clf_cpd} builds upon Algorithm~\ref{alg:cpd} by adding a binary classifier to compute the CPD score, and Algorithm~\ref{alg:gen_cpd} further builds on top by adding generative models for sample augmentation.}
  \label{fig:diagram}
\end{figure*}

The key property of generative modelling that our method exploits is the sampling from the learned distribution. Thus, the generated samples led the CPD score estimate to its limiting distribution. For a binary classifier from Algorithm~\ref{alg:clf_cpd}, to understand the behaviour of $\hat{D}$ with respect to the window size $n$, we rewrite $\hat{D}$ as:

\begin{equation}
\begin{split}
    \hat{D} &= \frac{1}{2n} \sum_{j=1}^{2n} \left[ \log \frac{P(y = 1 | x_j \in X_{ref})}{P(y = 0 | x_j \in X_{ref})} + \log \frac{P(y = 0 | x_j \in X_{test})}{P(y = 1 | x_j \in X_{test})}\right] \\
            &= \frac{1}{2n} \sum_{j=1}^{2n} \left[ \log \left( \frac{P(y = 1 | x_j \in X_{ref})}{P(y = 0 | x_j \in X_{ref})} \cdot \frac{P(y = 0 | x_j \in X_{test})}{P(y = 1 | x_j \in X_{test})} \right)\right] \\
            &= \frac{1}{2n} \sum_{j=1}^{2n} \left[ \log \left( \frac{P(y = 1 | x_j \in X_{ref})}{P(y = 0 | x_j \in X_{ref})} \Big/ \frac{P(y = 1 | x_j \in X_{test})}{P(y = 0 | x_j \in X_{test})} \right)\right] \\
            &= \frac{1}{2n} \sum_{j=1}^{2n} L,
\end{split}
\end{equation}
where $L$ is the sample log odds ratio. Sample log odds ratio $L$ is approximately normally distributed~\cite{log_or}:
\begin{equation}
    L \sim \mathcal{N}(\log(OR), s^2),
\end{equation}
where $\log(OR)$ is the true log odds ratio and the standard deviation is 
\begin{equation}\label{eq:d_std}
    s = \sqrt{\frac{1}{|TP|} + \frac{1}{|FN|} + \frac{1}{|FP|} + \frac{1}{|TN|}}
\end{equation}

based on the confusion matrix of the binary classifier of Algorithm~\ref{alg:clf_cpd}
\begin{table}[H]
\centering
\renewcommand{\arraystretch}{1.2} 
\begin{tabular}{cc|cc}
 & & \multicolumn{2}{c}{\textbf{Predicted}} \\
 & & $ \hat{y} = 1$ & $\hat{y} = 0$ \\
\cline{2-4}
\multirow{2}{*}{\rotatebox[origin=c]{90}{\textbf{Actual}}} 
 & $x \in X_{test}$ & True Positive (TP) & False Negative (FN) \\
 & $x \in X_{ref}$ & False Positive (FP) & True Negative (TN) \\
\cline{2-4}
\end{tabular}
\label{tab:confusion_matrix_binary}
\end{table}
Using the properties of Gaussian random variables, we get the following.
\begin{equation}\label{eq:score_dist}
    \hat{D} = \frac{1}{2n} \sum_{j=1}^{2n} L \sim \mathcal{N}\left( \log(OR), \frac{s^2}{2n} \right).
\end{equation}
The practical implication is that the variance of our CPD score estimator $\hat{D}$ is inversely proportional to the number of samples used. The proposed CPD algorithm, Algorithm ~\ref{alg:gen_cpd}, is based on Algorithm ~\ref{alg:clf_cpd} and adds a generative model to augment both the reference and test windows. By generating $m$ samples where $m \gg n$, we effectively reduce the variance of the CPD score at each time step $t$ by a factor of $n/m$. This 'variance dampening' is what allows the algorithm to distinguish true change points from stochastic noise, effectively increasing the signal-to-noise ratio of the CPD score. We also refer to Fig.~\ref{fig:diagram}. By its nested structure, the diagram shows exactly how Algorithm~\ref{alg:gen_cpd} builds on Algorithm~\ref{alg:clf_cpd}, which in turn builds on Algorithm~\ref{alg:cpd}.
\begin{algorithm}[ht!] 
\caption{Window-based binary classifier CPD with generative model}
\label{alg:gen_cpd}
\begin{algorithmic}[1]
\Require{CPD score $D$, data $X \in \mathbb{R}^{d \times N}$, window size $n$, generated window size $m$, binary classifier $Clf$, generative models $G_{ref}$ and $G_{test}$, peak threshold $\tau$} 

\State Initiate CPD score estimates: $\hat{D} \gets \varnothing$
\For{$t = n \dots N - n$ do}
    \State $X_{ref} \gets X_{t - n + 1 : t}$;
    \State $X_{test} \gets X_{t + 1 : t + n}$;
    \State Fit generative models: $G_{ref}(X_{ref}), G_{test}(X_{test})$;
    \State Generate: $\hat{X}_{ref} \gets G_{ref}(m), \hat{X}_{test} \gets G_{test}(m)$;
    \State $y_{ref} \gets 0, y_{test} \gets 1$;
    \State Fit classifier: $Clf ((\hat{X}_{ref}, y_{ref}), (\hat{X}_{test}, y_{test}))$;
    \State Compute CPD score at $t$: $d_t \gets D(Clf, \hat{X}_{ref}, \hat{X}_{test})$;
    \State Update CPD score estimates: $\hat{D} \gets \hat{D} \bigcup \{d_t\}$;
\EndFor
\State $\hat{T} \leftarrow \{t \mid d_t \text{ is a local maximum and } d_t > \tau\}$;
\State \Return $\hat{T}$.

\end{algorithmic}
\end{algorithm}

\section{Experiments}
\label{sec:exp}

In this section, we present experiments conducted to evaluate the performance of the proposed CPD algorithm. All experiments for all algorithms are conducted using the same values of shared parameters, such as, the window size. Table~\ref{tab:dataset_summary} summarises the experiments described in this section.

\vspace{-5mm}

\begin{table*}[htbp]
\centering
\begin{tabular}{lcccccc}
\toprule
\textbf{Dataset} & \textbf{Dim.} & \textbf{\#DS} & \textbf{Length} & \textbf{\#CPs}  & \textbf{Window size} & \textbf{Margin}\\
\midrule
Mean/Variance Jumps & $1$ & $10$ & $5000$ & $49$ & $50$ & $20$ \\
Covariance Jumps & $2$ & $10$ & $5000$ & $49$ & $50$ & $20$ \\
Bee Dance & $3$ & $6$ & $828 \pm 202$ & $20.0 \pm 4.4$ & $5$ & $10$  \\
HASC-2011 & $1$ & $10$ & $2938 \pm 806$ & $11.0 \pm 2.8$ & $40$ & $50$ \\
USC-HAD & $3$ & $14$ & $3394 \pm 863$ & $14.0 \pm 0.0$ & $70$ & $20$ \\
WISDM & $6$ & $20$ & $2188 \pm 82$ & $17.0 \pm 0.3$ & $60$ & $20$  \\
\bottomrule
\end{tabular}
\caption{Summary of datasets including dimensionality, number of time series (\#DS), mean $\pm$ standard deviation of lengths, mean $\pm$ standard deviation of the number of change points (\#CPs), window sizes used for window-based algorithms, and margins used for computing quality metrics.}
\label{tab:dataset_summary}
\end{table*}

\vspace{-15mm}

\subsection{Synthetic data sets}\label{sec:simdata}

Synthetic datasets allow for a fair comparison under controlled conditions for a specific type of change points. Following ~\cite{LIU201372} and ~\cite{chang2019kernel}, for each type of change point, we created 10 datasets. Each length $N=5000$ contain change points in random time steps $t_k : t_0 = 0, t_k = t_{k-1} + \lfloor \tau_k \rfloor$, where $\tau_k \overset{\hphantom{\text{i.i.d.}}}{\sim} \mathcal{N}(100, 10)$.
    
\textbf{Mean jumps}. Following ~\cite{deryck2021change}, we consider a one-dimensional autoregressive process (AR):

\begin{equation}\label{eq:ar}
    y_t = 0.6 y_{t-1} - 0.5 y_{t-2} + \varepsilon_t,
\end{equation}

where $\varepsilon_t \overset{\hphantom{\text{i.i.d.}}}{\sim} \mathcal{N}(\mu_t, \sigma^2_t)$ and $y_1 = y_2 = 0$. For mean jumps, we consider $\sigma_t = 1.5$ and the noise mean is as follows: \\
    \begin{equation}
        \mu_t = 
        \begin{cases}
            0 & 1 \leq t \leq t_1 \\
            \mu_{t_{k-1}} + \frac{k}{16} & t_{k-1} + 1 \leq t \leq t_k. 
        \end{cases}
    \end{equation}
    
\textbf{Variance jumps}. Analogically to mean jumps, we take the AR model, Eq.~\ref{eq:ar}. For variance jumps, we take the mean of the noise term as $\mu_t = 0$ and the standard deviation as follows: \\
    \begin{equation}
        \sigma_t = 
        \begin{cases}
            1 & \text{ if k is odd} \\
            \ln(e + \frac{k}{4}) & \text{ if k is even.}
        \end{cases}
    \end{equation}
        
\textbf{Covariance jumps}. Observations are individually and independently drawn from a two-dimensional normal distribution with mean vector $\mu = \left[ 0, 0 \right]^T$ and covariance matrix defined as follows: \\
    \begin{equation}
        \Sigma_t = 
        \begin{cases}
            \begin{bmatrix}
            1 & \delta \cdot k \\
            \delta \cdot k & 1
            \end{bmatrix} & \text{ if k is odd} \\
            \begin{bmatrix}
            1 & - \delta \cdot k \\
            - \delta \cdot k & 1
            \end{bmatrix} & \text{ if k is even,}
        \end{cases}
    \end{equation}
where $\delta = \lfloor \frac{c}{N} \rfloor$ and $c$ is the average number of observations between the change points. Thus, the absolute value of the covariance is at most $1$.

\subsection{Real data sets}
\label{sec:realdata}

In addition to the synthetic datasets, we tested the CPD algorithm on real-world datasets. They are inherently more complex and often contain subtle or gradual change points that challenge the detection capabilities of the algorithm. By testing these datasets, we assessed the effectiveness of the model under realistic conditions, where noise, non-stationarity, and unstructured data patterns may be present.

The \textbf{Bee Dance} dataset~\cite{bee-dance} are six records of bee movements, with change points occurring when bees change from one type of movement to another, such as directional changes in their waggle dance. Detecting these change points accurately demonstrates the sensitivity of the algorithm to natural, nonlinear behavioural shifts, which is valuable for applications in biology and behavioural science. The successful identification of change points in this dataset indicates that the CPD algorithm can adapt to complex and non-stationary motion patterns.
    
\textbf{HASC-2011}~\cite{10.1145/3123024.3124410}, \textbf{USC-HAD}~\cite{mi12:ubicomp-sagaware}, and \textbf{WISDM}~\cite{8835065} are three commonly used human-activity datasets with $10, 14$, and $20$ records respectively. Each data set contains sensor data that capture the movements of a participant, with change points representing the transitions between activities, for example, walking to running and back. In these datasets, change points are most commonly a combination of at least two change point types considered in the previous section, for example, simultaneous mean and covariance jumps. 

\begin{itemize}
    \item \textbf{HASC-2011}: Following ~\cite{LIU201372} and ~\cite{deryck2021change}, a three-dimensional accelerometer dataset was reduced to one dimension using the 2-norm, capturing transitions between different activity states.
    \item \textbf{USC-HAD}: Similar to HASC-2011, this dataset comprises three-dimensional accelerometer recordings that provide a multivariate view of activity changes.
    \item \textbf{WISDM}: A more complex data set with a six-dimensional time series from accelerometer and gyroscope sensors, allowing us to test the algorithm’s performance in a higher-dimensional setting.
\end{itemize}

\subsection{Quality Metrics}
\label{sec:quality}

Each of the CPD algorithms described above, namely Algorithms \ref{alg:cpd}, \ref{alg:clf_cpd}, and \ref{alg:gen_cpd}, makes a prediction by finding all relative maxima, that is, peaks in the CPD score. To compute the set of correctly identified change points, True Positive (TP), a set of non-negative threshold values is first defined up to the largest CPD score value. Then, for each threshold value, $\tau$, we filter out maxima that are lower than said threshold. Taking the remaining maxima as the predicted change points $\hat{t} \in \hat{T}$ and the set of ground truth change points $t^{*} \in T^*$ TP is defined as follows:

\begin{equation}
\begin{split}
    \text{TP} = \{ \hat{t}_i \in \hat{T} &\mid \exists t_j^* \in T^* : |\hat{t}_i - t_j^*| < M,\\ &\quad \hat{t}_i = \arg\min_{\hat{t} \in \hat{T}} \big( |\hat{t} - t_j^*|, \hat{t} \big) \}
\end{split}
\end{equation}

for some margin $M$. The margin $M$ (listed for each dataset in Table \ref{tab:dataset_summary}) defines the temporal tolerance for a correct detection. A predicted change point $\hat{t}_i$ is considered a True Positive only if it falls within $\pm M$ time steps of a ground truth point $t^*_j$. This margin accounts for the inherent delay in window-based detection, where the CPD score typically peaks only after the 'test window' has fully captured the new distribution. Note that all ground truths are paired with a prediction only ones. Thus, there is no overflow when computing TP. Then Precision and Recall can be computed as follows:

\begin{equation}
    \text{Precision} = \frac{|TP|}{|\hat{T}|},
\end{equation}

\begin{equation}
    \text{Recall} = \frac{|TP|}{|T^{*}|},
\end{equation}

where $|\cdot|$ denotes the cardinality of set. Thus, for each threshold value, $\tau$, we obtain precision and recall. As the primary CPD quality metric we use area under the precision recall curve (\textbf{PR AUC}). It summarises the performance of the algorithm over all threshold values, which allows us to estimate the overall performance of the algorithm while incorporating the intricacies of the CPD score and mitigating any minor computational error.

In contrast to other quality metrics commonly used in CPD literature, such as ROC AUC or F1, Covering, NAB, and RCPD \cite{bee-dance}, PR AUC accounts for the rare occurrence of change points and, instead of being a single-point evaluation, it is an integral metric over all CPD score values. PR AUC provides a more rigorous assessment of the algorithm's ability to maintain high precision (avoiding false alarms) while achieving high recall, which is the primary challenge in real-world CPD applications.

\subsection{Generative Models}
\label{sec:gms}

In Algorithm~\ref{alg:gen_cpd}, for each pair of reference and test windows, we used a separate generative model to learn the corresponding underlying distributions. In general, the choice of generative model should be motivated by experiment specifics. We use three distinct types of generative model: Gaussian Mixture Models (GMM), Variational Autoencoders (VAE), and Normalising Flows (NF).

GMM~\cite{GMM} is a classical statistical approach to generative modelling, where data are modelled as a mixture of multiple Gaussian distributions. By fitting GMMs to the reference and test windows, we generate synthetic samples that maintained key statistical characteristics, such as the mean and variance, of the original time-series data.

VAE~\cite{vae} is a class of deep generative models. An encoder neural network learns a compressed latent representation of data and a decoder neural network uses it to reconstruct the input through a probabilistic framework. VAEs offer a flexible way to model complex data distributions, making them suitable for high-dimensional or nonlinear time series where simple Gaussian assumptions may be insufficient.

NFs~\cite{nf} transform simple distributions, for example, Gaussian distributions, into complex distributions by applying a series of invertible transformations, each being a neural network. NFs are particularly effective for data augmentation in CPD tasks, in which the data distribution deviates significantly from normality. Using a series of transformations, NF-based models can learn highly flexible and detailed representations of time series data, making them suitable for applications that require precise alignment between synthetic and real data distributions.

\subsection{Data Augmentation Methods}
\label{sec:aug}

To validate the use of generative modelling, as control, we run Algorithm \ref{alg:gen_cpd} with Bootstrap and SMOTE instead of generative models.

Bootstrap~\cite{bootstrap} is a straightforward re-sampling strategy that randomly selects observations from the original dataset, effectively maintaining the sample's statistical properties without introducing new information. On the other hand, Synthetic Minority Oversampling TEchnique (SMOTE)~\cite{SMOTE} generates new data points using the nearest neighbours of each observation; although it creates new samples, it does not ensure the preservation of the statistical properties of the original sample.

Generative models transcend these limitations by offering a combination of the advantages of both bootstrapping and SMOTE. They not only preserve the statistical integrity of the sample, such as bootstrapping, but also generate new samples, similar to SMOTE. Furthermore, generative models have the potential to capture complex data distributions, providing richer augmentations that can enhance the performance of machine-learning models.

\subsection{Baseline Methods}\label{sec:base}

We compared our method with baselines that cover a range of various approaches to the problem and have shown a high quality of prediction. Statistical Window Sliding (Window) ~\cite{TRUONG2020107299}, Binary Segmentation (BinSeg) ~\cite{TRUONG2020107299}, and machine learning approaches include kernel learning with auxiliary deep generative models (KL-CPD) ~\cite{chang2019kernel}, autoencoders with a time-invariant representation (TIRE) ~\cite{deryck2021change}, and relative unconstrained least squares fitting (RuLSIF) ~\cite{LIU201372}. All baseline models were tuned using a rigorous grid search over their respective hyperparameter spaces (as defined in their original papers). To ensure a fair comparison, the total computational budget for tuning each baseline was equivalent to or greater than that allocated to our proposed method.

In addition, we consider Algorithm~\ref{alg:clf_cpd}, equipped either with Quadratic Discriminant Analysis (QDA) or Decision Tree (DT). Both algorithms used QDA and DT as the base classifiers. For simplicity, we consider the QDA and DT. If the experiment requires a more flexible classifier, for example, when data are high-dimensional, models such as neural networks can be considered ~\cite{deryck2021change}.

To demonstrate that the proposed approach is both consistent and adaptable, we consider Algorithm~\ref{alg:gen_cpd} as an extension of the QDA, DT, and Window algorithms. The latter was chosen to show that our approach is also viable with statistical methods, that is, it can be used as an extension of Algorithm~\ref{alg:cpd}.

In practical applications where ground truth is unavailable, window size $n$ can be determined via autocorrelation analysis or domain-specific knowledge of the minimum expected duration between events. For this comparative study, we fixed $n$ to a value representative of the dataset's scale to ensure all baselines were evaluated under identical temporal constraints, avoiding the 'tuning bias' that occurs when window sizes are optimized per-method.

\begin{center}
\begin{table*}[htbp]\centering\normalsize
\subfloat[PR AUC values for change point detection algorithms on synthetic datasets.]{\begin{tabular}{|l|r|r|r|}
\hline
CPD Algorithm    &  Mean jumps         &  Variance jumps     &  Covariance jumps \\
\hline

Binseg        &  $0.911 \pm 0.009$        &  $0.693 \pm 0.020$        &  $0.626 \pm 0.019$  \\

KL-CPD  (1)      &  $0.891 \pm 0.011$        &  $0.668 \pm 0.035$        &  $0.294 \pm 0.020$  \\

TIRE  (1)        &  $0.754 \pm 0.047$        &  $0.889 \pm 0.014$        &  $0.227 \pm 0.015$  \\

RuLSIF (2)       &  $0.943 \pm 0.011$        &  $0.894 \pm 0.014$        &  $0.797 \pm 0.017$  \\

\hline

QDA  (2)     &  $0.922 \pm 0.015$        &  $0.863 \pm 0.019$        &  $0.903 \pm 0.011$  \\

GMM QDA (3)       &  \cellcolor{yellow!25}$0.936 \pm 0.009$        &  \cellcolor{green!25}$\mathbf{0.897 \pm 0.007}$  &  \cellcolor{green!25}$\mathbf{0.917 \pm 0.008}$  \\

VAE QDA (3)       &  \cellcolor{green!25}$\mathbf{0.946 \pm 0.006}$      &  \cellcolor{red!25}$0.855 \pm 0.010$        &  \cellcolor{red!25}$0.887 \pm 0.017$  \\

NF QDA (3)        &  \cellcolor{yellow!25}$0.932 \pm 0.010$        &  \cellcolor{red!25}$0.856 \pm 0.012$        &  \cellcolor{red!25}$0.835 \pm 0.027$  \\

Bootstrap QDA (3) &  \cellcolor{yellow!25}$0.927 \pm 0.008$        &  \cellcolor{red!25}$0.858 \pm 0.012$        &  \cellcolor{green!25}$0.911 \pm 0.010$  \\

SMOTE QDA (3)     &  \cellcolor{yellow!25}$0.931 \pm 0.006$        &  \cellcolor{yellow!25}$0.888 \pm 0.015$        &  \cellcolor{green!25}$0.907 \pm 0.009$  \\

\hline

DT (2)      &  $0.855 \pm 0.017$        &  $0.742 \pm 0.036$        &  $0.770 \pm 0.020$  \\

GMM DT (3)    &  \cellcolor{green!25}$0.945 \pm 0.012$        &  \cellcolor{yellow!25}$0.875 \pm 0.007$        & \cellcolor{green!25}  $0.900 \pm 0.017$  \\

\hline

Window (1)     &  $0.942 \pm 0.005$        &  $0.861 \pm 0.014$        &  $0.784 \pm 0.011$  \\

GMM Window (3)   &  \cellcolor{green!25}$\mathbf{0.946 \pm 0.006}$ &  \cellcolor{yellow!25}$0.879 \pm 0.009$        & \cellcolor{green!25}  $0.828 \pm 0.014$  \\

\hline
\end{tabular}
}\qquad
\subfloat[PR AUC values for change point detection algorithms on real datasets.\vspace{5mm}]{\begin{tabular}{|l|r|r|r|r|}
\hline
CPD Algorithm       &  Bee Dance                &  HASC-2011                &  USC-HAD                  &  WISDM \\
\hline

Binseg        &  $0.474 \pm 0.051$        &  $0.776 \pm 0.041$        &  $0.712 \pm 0.028$        &  $0.882 \pm 0.025$  \\

KL-CPD (1)       &  $0.380 \pm 0.035$        &  $0.691 \pm 0.069$        &  $0.781 \pm 0.032$        &  $0.847 \pm 0.022$  \\

TIRE (1)         &  $0.483 \pm 0.045$        &  $0.745 \pm 0.049$        &  $0.803 \pm 0.030$        &  $0.719 \pm 0.043$  \\

RuLSIF (2)       &  $0.503 \pm 0.046$        &  $0.611 \pm 0.122$        &  $0.746 \pm 0.060$        &  $0.908 \pm 0.042$  \\

\hline

QDA (2)     &  $0.647 \pm 0.016$        &  $0.763 \pm 0.096$        &  $0.849 \pm 0.044$        &  $0.928 \pm 0.032$  \\

GMM QDA (3)      &  \cellcolor{green!25}$\mathbf{0.658 \pm 0.015}$  &  \cellcolor{green!25}$0.780 \pm 0.056$        &  \cellcolor{green!25}$0.865 \pm 0.024$        &  \cellcolor{green!25}$\mathbf{0.954 \pm 0.014}$  \\

VAE QDA (3)       &  \cellcolor{red!25}$0.506 \pm 0.021$        &  \cellcolor{red!25}$0.756 \pm 0.059$        &  \cellcolor{red!25}$0.585 \pm 0.063$        &  \cellcolor{red!25}$0.910 \pm 0.017$  \\

NF QDA (3)        &  \cellcolor{red!25}$0.572 \pm 0.037$        &  \cellcolor{red!25}$0.745 \pm 0.054$        &  \cellcolor{red!25}$0.381 \pm 0.070$        &  \cellcolor{red!25}$0.718 \pm 0.070$  \\

Bootstrap QDA (3) &  \cellcolor{red!25}$0.592 \pm 0.036$        &  \cellcolor{red!25}$0.761 \pm 0.060$        &  \cellcolor{green!25}$\mathbf{0.903 \pm 0.022}$    &  \cellcolor{red!25}$0.899 \pm 0.014$  \\

SMOTE QDA (3)     &  \cellcolor{red!25}$0.645 \pm 0.011$        &  \cellcolor{red!25}$0.754 \pm 0.058$        &  \cellcolor{green!25}$0.889 \pm 0.024$        &  \cellcolor{red!25}$0.909 \pm 0.013$  \\

\hline

DT (2)      &  $0.629 \pm 0.019$        &  $0.773 \pm 0.082$        &  $0.775 \pm 0.040$        &  $0.895 \pm 0.052$  \\

GMM DT (3)    &  \cellcolor{green!25}$\mathbf{0.658 \pm 0.010}$        &  \cellcolor{green!25}$\mathbf{0.796 \pm 0.054}$   &  \cellcolor{green!25}$0.832 \pm 0.024$        &  \cellcolor{yellow!25}$0.906 \pm 0.020$  \\

\hline

Window (1) &  $0.336 \pm 0.021$        &  $0.752 \pm 0.118$        &  $0.827 \pm 0.037$        &  $0.936 \pm 0.036$  \\

GMM Window (3)   &  \cellcolor{yellow!25}$0.342 \pm 0.018$        &  \cellcolor{green!25}$0.791 \pm 0.060$        &  \cellcolor{green!25}$0.843 \pm 0.017$        &  \cellcolor{green!25}$0.939\pm 0.015$  \\

\hline
\end{tabular}}
\caption{Mean and standard deviation of PR AUC across synthetic and real datasets. The best result per dataset is in bold. \textbf{Red}: Performance degraded compared to base method. \textbf{Yellow}: Performance improved over base but not best overall. \textbf{Green}: Performance improved over base and outperformed all baselines. Numbers $1, 2, 3$ in the index column indicated whether the CPD Algorithm follows Algorithm \ref{alg:cpd}, \ref{alg:clf_cpd}, or the proposed Algorithm \ref{alg:gen_cpd}, respectively.}
\label{table:sim}
\end{table*}
\end{center}


\section{Results}
\label{sec:results}

Table~\ref{table:sim} presents the main results of this study. Among the baseline methods, different models yielded the best score in separate datasets. RuLSIF, for example, outperforms all other baselines on synthetic data and Bee Dance, and vice versa for the window algorithm, and Binseg shows the best result on HASC. Deep learning methods, namely TIRE and KL-CPD go back and forth in terms of PR AUC. Note that both algorithms failed in the covariance jumps. The base algorithms, that is, QDA and DT, go back and forth with respect to the baseline methods, with QDA often coming on top, which is expected.

The results for QDA are studied more extensively than those for DT and Window in the context of the proposed approach than because the base algorithm yields better results across the board. Improving these results is challenging.

Table~\ref{table:sim} shows that adding a GMM consistently improves the performance of all base algorithms. More often than not, the extended algorithm outperforms all baselines (in green), with GMM QDA returning the best overall result. Other generative models and data augmentation methods worsen the result of their base counterpart (in red).

Both bootstrap and SMOTE can yield excellent results, and they are better than their GMM counterpart. However, the consistency of these results remains questionable.

The poor performance of the deep generative models can be attributed to the lack of data to properly fit a neural network and the simplicity of both architectures. However, advancing either the architecture of the hyperparameters or any other further makes the models computationally infeasible \cite{wang2024sequential}. Between the two deep GMs, the VAE has an edge over the NF and shows potential to improve the base CPD algorithm. Although there are some exceptions, it is crucial to note that these observations are generally true. This indicates that the behaviour described above is systematic.

This effect should be further studied. Consider algorithms with a GMM and their corresponding base counterparts. For each pair, let their PR AUCs be $\mu_{GMM} \pm \sigma_{GMM}$ and $\mu_{base} \pm \sigma_{Base}$ for the GMM and base algorithm, respectively. Then for each experiment, the standardised mean difference was calculated as follows:

\begin{equation}
    SMD = \frac{|\mu_{GMM} - \mu_{Base}|}{\sqrt{\sigma_{GMM}^2 + \sigma_{Base}^2}}.
\end{equation}

Both QDA and Window produce very high results, and thus their average $SMD$ across all datasets is approximately $0.8$, which indicates a moderate improvement. However, the average $SMD$ of DT is $2.53$; that is, GMM has an immense impact on the performance of DT.

\section{Discussion}
\label{sec:discussion}

This study proposes enhancing CPD through generative model data augmentation rather than through increasingly complex algorithms. GMMs provide an ideal middle ground, combining the efficiency of simple statistical methods with the density-learning capabilities of deep generative models. While high-capacity models like VAEs and NFs excel with image data, they often fail in small-window CPD regimes—such as the Bee Dance dataset—where they overfit to noise and "hallucinate" synthetic data due to an excess of trainable parameters relative to observations. In these contexts, GMMs act as superior high-bias, low-variance regularisers that avoid the performance degradation seen in deep models. Furthermore, because offline CPD requires retraining a model at every sliding window time-step, GMMs offer a decisive computational advantage; the EM algorithm is orders of magnitude faster than backpropagation, making GMM-based augmentation the only feasible choice for real-time applications or long time-series data.

\section{Conclusion}
\label{sec:conclusion}

In this work, we studied the effect of generative modelling for data augmentation for change point detection. We demonstrate that a simple generative model, such as a Gaussian Mixture Model, can significantly improve the performance of a change point detection algorithm. This approach outperforms all the selected baselines that include a wide range of approaches. The use of a deep generative model or other data augmentation techniques such as Bootstrap or SMOTE is justified. However, extensive experimentation shows that these alternatives often worsen the performance of the base algorithm.

\section*{FUNDING}

The article was prepared within the framework of the project “Mirror Laboratories” HSE University. The computation for this research was performed using the computational resources of HPC facilities at HSE University~\cite{Kostenetskiy_2021}.

\section*{ETHICS APPROVAL AND CONSENT TO PARTICIPATE}

This work does not contain any studies involving human and animal subjects.

\section*{CONFLICT OF INTEREST}
The authors of this work declare that they have no conflicts of interest.

\bibliographystyle{unsrt}
\bibliography{mybibfile}

\end{document}